\documentclass{article}

\usepackage[dvips]{graphicx}
\usepackage{inputenc}
\usepackage{amssymb,amsmath,array}
\usepackage{hyperref}
\usepackage[capitalise]{cleveref}
\usepackage{subcaption}
\usepackage{booktabs}
\usepackage{algorithm}
\usepackage{algorithmic}
\usepackage{wrapfig}
\usepackage{url}

\usepackage{todonotes}
\usepackage{multirow}
\newcommand{\D}{\mathcal{D}}

\newcommand{\T}{\mathcal{T}}

\renewcommand{\P}{\mathbb{P}}

\newcommand{\orcidID}[2]{#1\footnotemark[0]\footnotetext[0]{#1: \url{#2}}}
\newcommand{\authorOID}{\orcidID{Fabian Hinder}{https://orcid.org/0000-0002-1199-4085}, \orcidID{Valerie Vaquet}{https://orcid.org/0000-0001-7659-857X}, \orcidID{Johannes Brinkrolf}{https://orcid.org/0000-0002-0032-7623}, and \orcidID{Barbara Hammer}{https://orcid.org/0000-0002-0935-5591}}

\begin{document}
\title{Drift Localization using Conformal Predictions\footnote{Extension of paper \cite{thisEsann} published at the 34th European Symposium on Artificial Neural Networks, Computational Intelligence and Machine Learning  (ESANN), 2026.}}

\author{\small 
\renewcommand{\orcidID}[2]{##1}
\authorOID\thanks{Funding in the scope of the BMFTR project KI Akademie OWL under grant agreement No. 16IS24057A and the ERC Synergy Grant ``Water-Futures'' No. 951424 is gratefully acknowledged.}
\vspace{.3cm}\\
\small Bielefeld University---Faculty of Technology \\
\small Inspiration 1, 33619 Bielefeld---Germany \\
\small \texttt{\{fhinder,vvaquet,jbrinkro,bhammer\}@techfak.uni-bielefeld.de}
}
\maketitle

\footnotetext[0]{\!\!\!\colorbox{white}{\color{white}${}^0$}\!\!\!\!\!\!\!\!\!\!\!\!\!\!\!\!\!\!\! \renewcommand{\orcidID}[2]{\includegraphics[height=0.8\baselineskip]{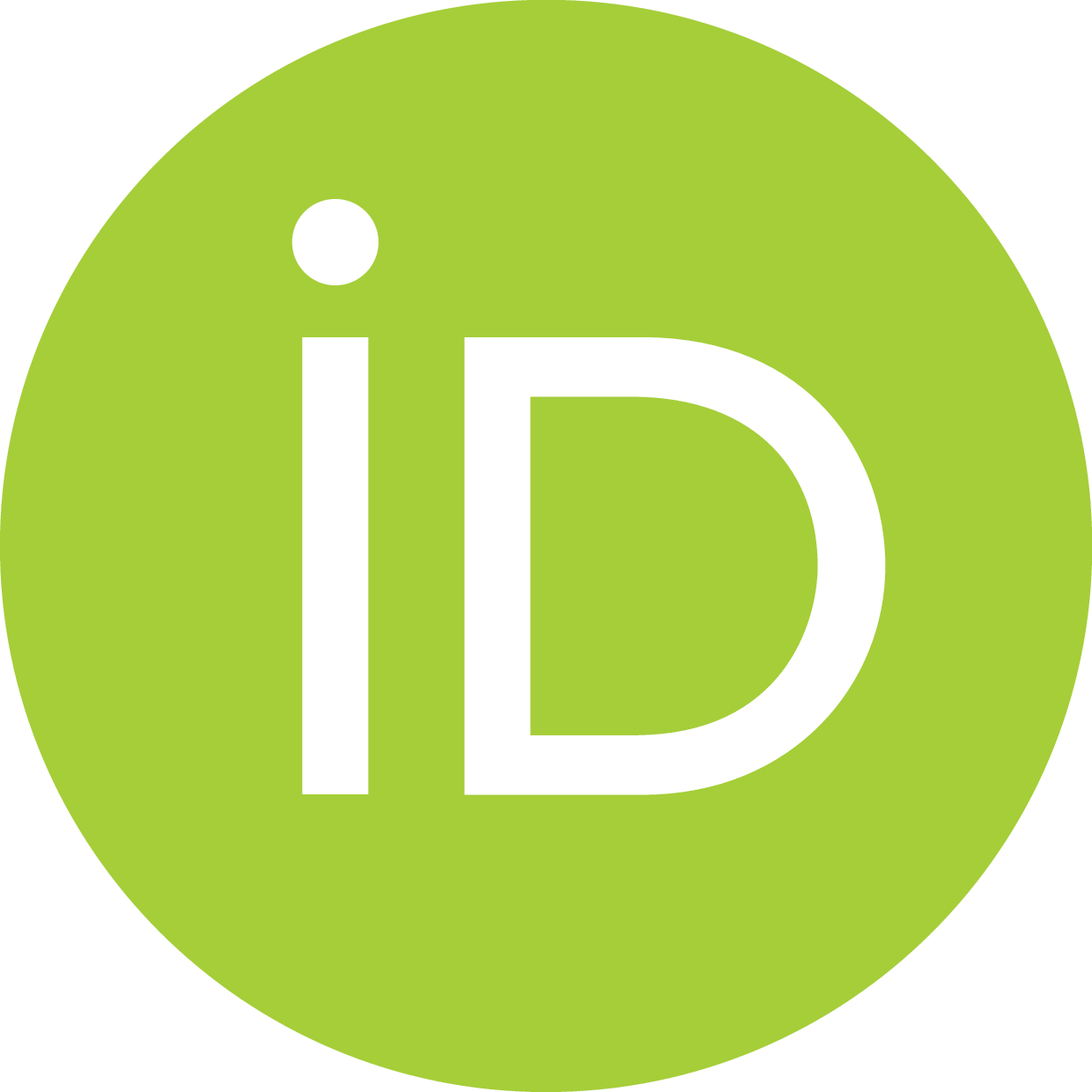} #1: \url{#2}}\authorOID}

\begin{abstract}
Concept drift---the change of the distribution over time---poses significant challenges for learning systems and is of central interest for monitoring.  Understanding drift is thus paramount, and drift localization---determining which samples are affected by the drift---is essential. While several approaches exist, most rely on local testing schemes, which tend to fail in high-dimensional, low-signal settings. In this work, we consider a fundamentally different approach based on conformal predictions. We discuss and show the shortcomings of common approaches and demonstrate the performance of our approach on state-of-the-art image datasets.
\end{abstract}

\section{Introduction}
\label{sec:intro}

In the classical batch setting of machine learning, data are assumed to originate from a single source and to be independently and identically distributed (i.i.d.). A relevant extension is to allow several different data distributions. Examples include transfer learning, where source and target domains differ; federated learning, where models are trained on data collected across distributed locations; and stream learning, where each time step may follow a different distribution~\cite{lu2018learning,oneortwo}.
In all these settings, the fact that the distributions may differ, a phenomenon referred to as \emph{concept drift}~\cite{lu2018learning} or drift for short, plays a central role. In stream learning and system monitoring, this relates to changes over time; in transfer learning, to the differences between source and target distributions, etc.

Gaining understanding of the drift is imperative~\cite{oneortwo,mbde}.
Recent work on \emph{model-based drift explanations}~\cite{mbde} has enabled the usage of generic explainable AI (XAI) techniques to obtain such insights. A key step of the approach is the identification of the affected samples, a task referred to as \emph{drift localization}~\cite{oneortwo,hinder2022}. While there are some methods addressing drift localization~\cite{oneortwo,lu2018learning,kdqtree,ldd,hinder2022,eikmeans,quadtree}, they mainly rely on local statistical testing, which tends to fail on high-dimensional data, such as streams of images.
In this work, we propose an alternative localization scheme by applying conformal prediction to the underlying idea of model-based drift localization~\cite{hinder2022}. We first summarize the setup, discuss the shortcomings of the related work, and recap conformal prediction (\cref{sec:relwork}). We then present our novel methodology in \cref{sec:method} and evaluate it on two established and one novel image stream in \cref{sec:exp}, before we conclude this work in \cref{sec:concl}.

\section{Related Work\label{sec:relwork}}
\subsection{Concept Drift and Drift Localization}
In this work, we mainly consider concept drift in the context of stream learning and system monitoring; however, the definitions presented below also extend to the other drifting scenarios discussed in the introduction. In stream learning, we model an infinite sequence of independent observations $X_1, X_2, \dots$, each drawn from a potentially different distribution $X_i \sim \D_i$~\cite{lu2018learning}. Drift occurs when $\D_i \neq \D_j$ for some $i, j$. A fully probabilistic framework was proposed in \cite{hinder2020}, augmenting each sample $X_i$ with an observation timestamp $T_i$. In this formulation, concept drift is equivalent to statistical dependence between data $X$ and time $T$. 

Based on this idea, drift localization---the task of finding all samples affected by the drift---can be formalized as the local and global temporal distribution differing~\cite{hinder2022}, i.e., $L = \{x \::\: \P_{T \mid X = x} \neq \P_T\}$. It was shown in \cite[Thm.~1 and~2]{hinder2022} that for finite time points, e.g., ``before drift'' and ``after drift'', this set exists as the unique solution with certain properties and can be approximated using estimators. 
Since most algorithms detect drift between time windows~\cite{oneortwo,lu2018learning}, this justifies many approaches and reduces drift localization to a probabilistic binary classification problem and a statistical test to assess the severity of the mismatch between $\hat{\P}_{T \mid X = x}$ and $\hat{\P}_T$ relating to the data-point-wise $H_0$-hypothesis ``$x$ is not drifting'' \cite{hinder2022}. 

\subsection{Localization Methods and Their Shortcomings}

There are a few approaches for drift localization. Nearly all of them are based on comparing the time distribution of local groups of points with the global reference, thus aligning with the considerations of \cite{hinder2022}. Those groups are most commonly formed unsupervised, i.e., without taking the time point into account. Classical $kdq$-trees~\cite{kdqtree} and quad-trees~\cite{quadtree} recursively partition the data space along coordinate axes; other approaches use $k$-means variants~\cite{eikmeans}. Besides those partition-based, there exist $k$-neighborhood-based approaches like LDD-DIS~\cite{ldd}. 
The model-based approach, introduced in~\cite{hinder2022}, differs from those in so far as they use the time information to choose a better-suited grouping, i.e., by training a decision tree predicting $T$ based on $X$, yet, due to overfitting, data points used to construct the grouping cannot be used for analysis. Furthermore, they suggest a purely heuristic approach based on random forests.
While those approaches differ in what statistic and normalization technique they use, all explore the idea of local and global temporal differences.

This induces a triad-off problem: using no ($kdq$-tree, LDD-DIS, etc.) or only a few temporal information, the obtained grouping is sub-optimal, leading to inaccurate estimates $\hat{\P}_{T \mid x \in G}$; employing much to improve grouping, we are left with little data for the testing, resulting in low per-group test power---as the tests are performed separately on each group, even moderate test-set sizes and group numbers lead to comparably small per-group sample sizes. Both effects lead to an overall low testing power. We thus aim for a technique that allows for a global variance analysis. 

\subsection{Conformal Prediction}
When training a classifier, the target is usually to minimize the overall expected error. However, such a scheme does not give any guarantees about the correctness of a single prediction. While probabilistic classification improves on this situation in theory by providing class probabilities and thus a measure of uncertainty, models usually over- or underfit, leading to sub-optimal assessments. 

Conformal prediction constitutes an alternative scheme returning a set of potential classes $F(x) \subset \mathcal{C}$. This allows one to ensure that the correct class is in the set with arbitrary high certainty, i.e., $\P[Y \in F(X)] \geq \alpha$ for a conformal model $F$. Commonly, one minimizes the expected number of predicted classes.

The most common way to create a conformal model is to wrap a class-wise scoring function, for instance, a probabilistic classifier, into a calibrated model. This calibrated model must be trained on data that has not been used before, yet, since it only processes simple one-dimensional scores, the necessary calibration set can be comparably small. The usage of conformal $p$-values allows choosing $\alpha$ after training calibration, i.e., construct $p_y(x)$ so that $\P[Y \not\in \{y \::\: p_y(X) > \alpha\}] > 1-\alpha$ holds for all $\alpha$. 

\section{Conformal prediction for drift localization\label{sec:method}}
Current state-of-the-art drift localization techniques rely on local statistical testing and either employ unsupervised grouping methods---which are not performant when considering high-dimensional data---, or face a trade-off problem between optimizing grouping quality and test stability. In this work, we propose to leverage conformal prediction to obtain a global testing scheme. In the following, we first describe the general idea and then the algorithmic details. 

As discussed before, \cite{hinder2022} showed that a point $x$ is non-drifting if and only if the conditional entropy $H(T \mid X = x)$ is maximal, assuming a finite time domain and uniform temporal distribution, i.e., if we are maximally uncertain about the observation time point.
Specifically, in \cite{hinder2022}, this is used as a test statistic and then normalized using a permutation scheme.

\begin{figure}
  \hfill
  \begin{minipage}{0.45\textwidth}\centering
  \includegraphics[width=0.95\textwidth]{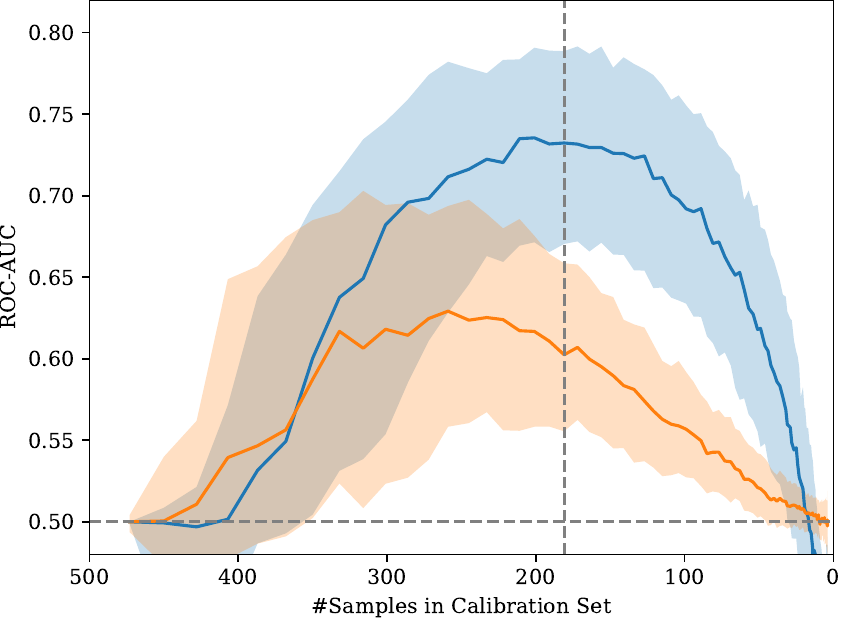}
  \subcaption{Size of grouping test/calibration \\ \;}
  \end{minipage}
  \hfill
  \begin{minipage}{0.45\textwidth}\centering
  \includegraphics[width=0.95\textwidth]{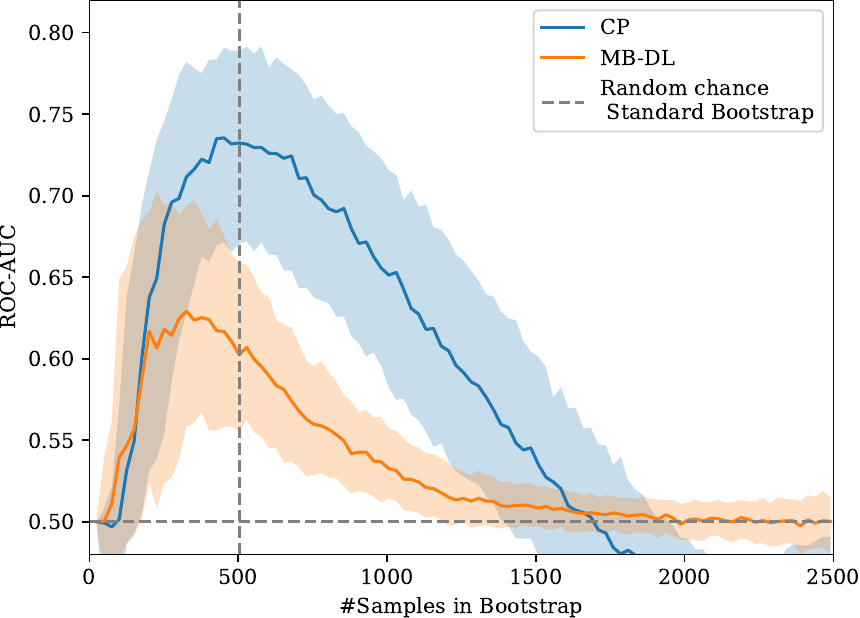}
  \subcaption{Size of over-/under \\ sampled bootstrap}
  \end{minipage}
  \hfill
  \caption{Effect of training-calibration/test-split on performance --- median and $25\%-75\%$-quantile. Decision tree, Fish-head dataset, 500 samples, 98 drifting. Plots differ in scaling of $x$-axis only (test/calibration set size reduces as size of bootstrap increases) \label{fig:splitsize}}
\end{figure}

To approach this using conformal predictions, recall that if $c \not\in F(x)$ then we can be very certain that the correct class is not $c$. Thus, following the ideas of \cite{hinder2022}, we can reject the hypothesis that $x$ is non-drifting if we can exclude one time point with certainty, i.e., $F(x) \neq \T$. By expressing $F_\alpha$ using conformal $p$-values, i.e., $F_\alpha(x) = \{y \in \mathcal{C} \::\: p_y(x) \geq \alpha\}$, we can reject $H_0$ if the minimal conformal score is smaller $\alpha$, i.e., we obtain the $p$-value $p_\text{loc}(X) = \min_y p_y(X)$. 
Using a probabilistic classifier as a class-wise scoring function, this relates to the probability of a large deviation of the class-probability from the global probability despite the sample being non-drifting, which is again in line with the considerations of \cite{hinder2022}. 

Using conformal prediction has various advantages compared to the local testing scheme. While conformal prediction still requires a calibration set, since this is only needed to calibrate a one-dimensional signal, it can be chosen much smaller while still offering good performance, allowing us to use more samples for model training. We display this effect in \cref{fig:splitsize}. 

Another advantage is that we are no longer limited to specific models. Commonly used local tests require the models to induce some kind of grouping. This is not the case for conformal prediction, thus making it compatible with any scoring function. This not only makes the method more compatible with the usage of supervised trained models but also allows a much larger pool of potential models to choose from.

\begin{algorithm}[!t]
\caption{Conformal Prediction for Drift Localization\label{alg}}
\begin{algorithmic}[1]
\footnotesize
\STATE \textbf{Input:} $(x_i,y_i)_{i=1}^n$, $y_i \in \mathcal{C}$; $n_{\mathrm{boot}}\in\mathbb{N}$
\STATE \textbf{Output:} $(p_i)_{i=1}^n$

\STATE $P_i \gets  [\,]$ \quad for all $i$

\FOR{$t = 1,\dots,n_{\mathrm{boot}}$}
    \STATE $(I_{\mathrm{in}}, I_{\mathrm{oob}}) \gets \textsc{SampleBootstrap}(\{1,\dots,n\})$
    \STATE $\theta \gets \textsc{TrainModel}(X_{I_{\mathrm{in}}}, y_{I_{\mathrm{in}}})$

    \FOR{$i \in I_{\mathrm{oob}}$}
        \STATE $P_i \gets P_i + \displaystyle\min_{c\in\mathcal{C}} \left[
            \frac{
        1 + \sum\limits_{k\in I_{\mathrm{oob}} \setminus \{i\}\::\: y_k = c} \mathbf{1}[f(c\mid x_k,\theta)\leq f(c\mid x_i,\theta)]
    }{
        1 + |\{k \in I_{\mathrm{oob}} \setminus \{i\} \::\: y_k = c \}|
    }
                \right]$ 
    \ENDFOR
\ENDFOR

\STATE $p_i \gets \mathrm{median}(P_i)$ \quad for all $i$
\STATE \textbf{Return} $(p_i)_{i=1}^n$
\end{algorithmic}
\end{algorithm}

The main hurdle for translating our considerations into an algorithm is the need for a calibration set to perform conformal predictions. We propose using bootstrapping as the out-of-bag samples constitute a natural calibration set of decent size, even when oversampling. The model is thus trained on the in-bag samples and then calibrated using the out-of-bag samples. Using the leave-one-out calibrated models, we assign $p$-values to the out-of-bag samples using the minimal conformal $p$-value---an out-of-training-sample extension is then obtained by using the full conformal model. 
To combine the resulting $p$-values across several bootstraps, we suggest using a median, which is equivalent to an ensemble of tests: we reject $H_0$ at level $\alpha$ if the majority of bootstraps lead to a rejection at level $\alpha$. 
The overall scheme is presented in \cref{alg}.

\section{Experiments\label{sec:exp}}
We are evaluating the proposed conformal-prediction-based approach on streams of images. More precisely, we rely on the Fashion-MNIST~\cite{FashionMNIST} dataset and the No ImageNet Class Objects (NINCO)~\cite{NINCO}. We follow the experimental setup by~\cite{hinder2022}, randomly selecting one class for non-drifting, drifting before, and drifting after.
While we use the Fashion-MNIST without further preprocessing, for NINCO, we use deep embeddings (DINOv2 ViT-S/14~\cite{oquab2023dinov2}).\footnote{DINOv2 (released April 2023) was trained on large-scale image data, whereas NINCO (released June 2023) is a curated out-of-distribution benchmark. Given NINCO's later release and evaluation-focused design, its inclusion in DINOv2's training data is unlikely, though this cannot be confirmed definitively.}

\paragraph{Fish-Head Dataset}
While this is a simple way to obtain drifting image data streams, we propose to also consider a more subtle drifting scenario where we do not model drift by switching up classes, but induce a more nuanced and realistic type of drift. For our novel benchmark \emph{Fish-Head data stream}, we rely on the ImageNet~\cite{imagenet} subset ``ImageNette'' consisting of the classes ``tench,'' ``English springer,'' ``cassette player,'' ``chain saw,'' ``church,'' ``French horn,'' ``garbage truck,'' ``gas pump,'' ``golf ball,'' and ``parachute.'' We manually split the class ``tench'' into the two sub-classes where the fish head is turned towards the left or right side, respectively; these are the drifting samples, all other samples are non-drifting. For our analysis, we again use a DINOv2 ViT-S/14~\cite{oquab2023dinov2} model to embed the images. A simple analysis shows that non-drifting, drifting before, and drifting after are lineally separable in the embedding.

\begin{figure}
    \centering
    \includegraphics[width=7cm,height=3cm]{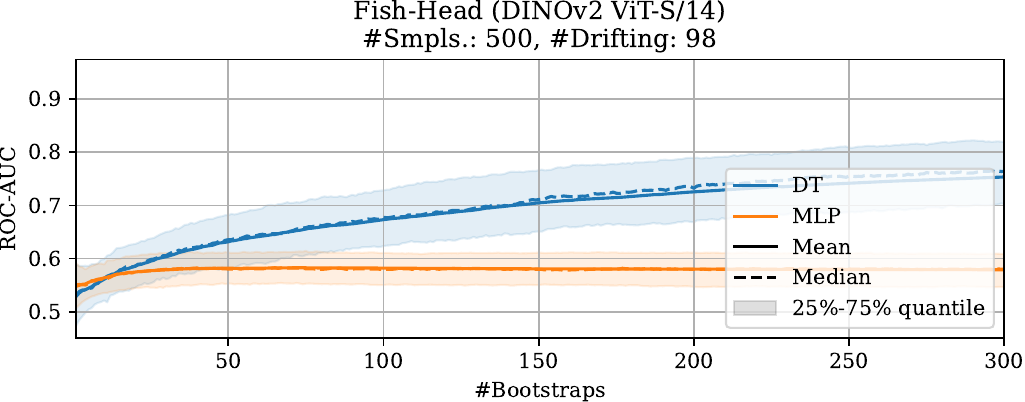}
    \caption{Effect of number of bootstraps on ROC-AUC. Figure shows aggregation of 500 runs on Fish-Head dataset.\label{fig:bootstrap}}
\end{figure}
\paragraph{Effect of Bootstraps}
If evaluating via ROC-AUC scores, our method has two remaining ``parameters:'' the inner model used to perform the prediction and the number of bootstraps. Here, our focus will be on the number of bootstraps. The results are presented in \cref{fig:bootstrap}. 

For our evaluation, we only take those data points into account that are assigned a value, i.e., those that are at least once out-of-bag and hence assigned a value via CP. The remaining points are not taken into account when forming the mean/median/quantile. This procedure is valid, as for more than 20 bootstraps $\geq 95\%$ all points are assigned a value. 

\begin{figure}
    \centering
    \includegraphics[width=\textwidth]{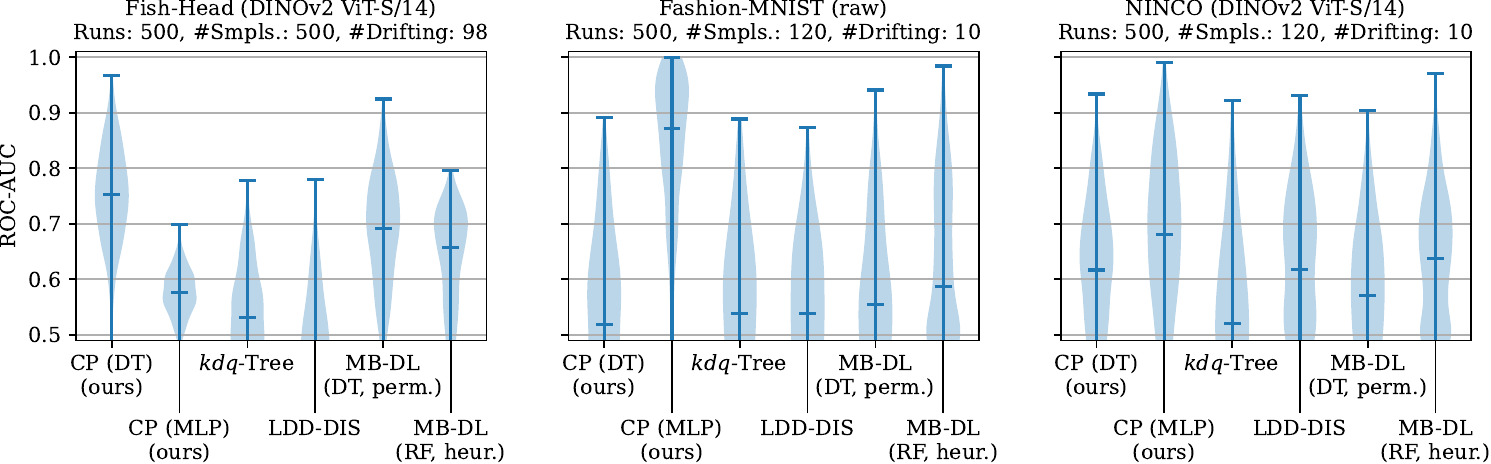}
    \caption{Experimental results. ROC-AUC (500 runs) for various drift localizes using windows of 250/60 samples with 98/10 drifting samples (in total).\label{fig:results}}
\end{figure}

Considering the actual results (\cref{fig:bootstrap}), we observe that for a small number of bootstraps the increase is significant, while for larger numbers it tends to plateau out. This effect is more visible for the MLP, which seems to reach its asymptotic state at about 30 samples. For DT, the number larger than 300---considering $5{,}000$ bootstraps shows the value stabilizes at a median ROC-AUC of $\approx 90\%$. However, using an even larger number of bootstraps further increases the computation time too significantly compared to that of other methods. In the final evaluation below (see \cref{fig:results}), where we only use 100 bootstraps, similar to \cite{hinder2022} with decision trees or inside the random forest.
Overall, we observe that the number of bootstraps increases the performance, which stabilizes at a sufficient, model specific size. 

\paragraph{Evaluation of Method}
To evaluate our method, we follow the same procedure as~\cite{oneortwo}\footnote{See \url{https://github.com/FabianHinder/Advanced-Drift-Localization} for the code}. For the simpler NINCO and Fashion-MNIST datasets, we use $2\times 60$ samples with 10 drifting, for the Fish-Head dataset, we used $2\times 250$ samples with 98 drifting; here, the increase was necessary as otherwise no method yielded results above random chance. Besides the novel conformal-prediction-based approach (CP) using decision trees (DT) and MLPs as models, we evaluate the model-based approach~(MB-DL; \cite{hinder2022}) using permutations and decision trees on bootstraps (DT, perm.) similar to \cref{alg} and heuristic approach based on random forests (RF, heur.), LDD-DIS~\cite{ldd}, and $kdq$-trees~\cite{kdqtree}. Following \cite{oneortwo}, we use the ROC-AUC as a score. We repeat each experiment 500 times. 

The results are shown in \cref{fig:results}. We observe that for both NINCO and Fashion-MNIST, the proposed conformal-prediction-based localization outperforms the related work when using MLPs. When using decision trees, the conformal prediction approach outperforms the permutation-based approach (MB-DL; DT) on NINCO; on Fashion-MNIST the method performs comparably poorly.  Considering the new Fish-Head benchmark, we observe that our method, realized with decision trees, outperforms the other methods. In this setting, the model-based decision trees outperform our MLP-based version and the random forest heuristic. The latter is interesting as the random-forest-based heuristic usually performs better than the actual permutation-based approach. This shows that the larger pool of potential base models is an important aspect of our conformal approach. Overall, the new data benchmark and the NINCO dataset seem to be more challenging tasks, making them suitable for further evaluation.

\section{Conclusion and Future Work\label{sec:concl}}

In this work, we proposed a novel strategy for drift localization, replacing the state-of-the-art local testing in data segments with a conformal-prediction-based global strategy allowing for a larger model pool while providing formal guarantees. We experimentally showed the advantage of the proposed methodology on established image data streams, showing that those can essentially be solved by the presented method. Besides, we proposed a novel image stream benchmark containing more subtle drift. 
As our experiments show, this benchmark is far more challenging, requiring a significantly larger number of samples to ensure results better than random chance. Further investigation and development that is better suited for small-sample size setups, and in particular, work in the case of only few drifting samples, is subject to future work. Furthermore, investigating the relevance of the used deep embedding seems to be a relevant consideration.

\begin{footnotesize}

\bibliographystyle{unsrt}
\bibliography{bibliography.bib}

\end{footnotesize}

\end{document}